\def\year{2022}\relax
\documentclass[letterpaper]{article} % DO NOT CHANGE THIS
\usepackage{aaai21}  % DO NOT CHANGE THIS
\usepackage{times}  % DO NOT CHANGE THIS
\usepackage{helvet} % DO NOT CHANGE THIS
\usepackage{courier}  % DO NOT CHANGE THIS
\usepackage[hyphens]{url}  % DO NOT CHANGE THIS
\usepackage{graphicx} % DO NOT CHANGE THIS
\usepackage{natbib}  % DO NOT CHANGE THIS AND DO NOT ADD ANY OPTIONS TO IT
\usepackage{caption} % DO NOT CHANGE THIS AND DO NOT ADD ANY OPTIONS TO IT
\usepackage{mathtools}
\usepackage[utf8]{inputenc}
\usepackage{hyperref}
\usepackage{booktabs}
\usepackage[many]{tcolorbox}
\newtcolorbox{post_box}[3][]{%
boxsep=3pt,left=2pt,right=2pt,bottom=3pt,
width=\columnwidth,
boxrule=1pt,
colbacktitle=white,coltitle=black,
boxed title style={size=normal,colframe=white,boxrule=0pt}, 
interior style={white},
enhanced,
float,
fonttitle=\scshape,
title=Example~\thetcbcounter: #2,
#1
}

\usepackage{amsthm}
\usepackage{amsmath,amsfonts}
\usepackage{amsmath}
\DeclareMathOperator*{\softmax}{softmax}

\DeclareMathAlphabet{\mathsl}{OT1}{ptm}{m}{sl}

\usepackage{xspace}

\newcommand{\citepos}[1]{\citeauthor{#1}'s \citeyearpar{#1}}

\newcommand{\ifsubmit}[1]{}

\newcommand{\be}{\begin{itemize}}
\newcommand{\ee}{\end{itemize}}
\newcommand{\bn}{\begin{enumerate}}
\newcommand{\en}{\end{enumerate}}
\newcommand{\bc}{\begin{center}}
\newcommand{\ec}{\end{center}}
\newcommand{\bl}{\begin{flushleft}}
\newcommand{\el}{\end{flushleft}}
\newcommand{\beq}{\begin{equation}}
\newcommand{\eeq}{\end{equation}}
\newcommand{\bq}{\begin{quote}}
\newcommand{\eq}{\end{quote}}

\newcommand{\bmp}{\begin{minipage}}
\newcommand{\emp}{\end{minipage}}

\DeclareMathAlphabet{\mathsl}{OT1}{ptm}{m}{sl}
\usepackage{xcolor}
\usepackage{siunitx}
\usepackage[np]{numprint}
\usepackage{multirow}
\npthousandsep{,}
\newcolumntype{T}{>{\tiny}l} % define a new column type for \tiny
\newcolumntype{H}{>{\Huge}l} % define a new column type for \Huge
\usepackage[inline]{enumitem}
\setlist[description]{leftmargin=1em}
\setlist[itemize]{leftmargin=1em}
\setlist[enumerate]{leftmargin=1.5em}

\usepackage{tikz}
\usepackage{standalone}
\usepackage{tikz-qtree}

\usepackage{soul}

\definecolor{revision}{rgb}{0.29, 0.0, 0.51}

\usepackage[normalem]{ulem}

\title{Conversation Modeling to Predict Derailment}
\author{
    Jiaqing Yuan,
    Munindar P. Singh
}
\affiliations{
    \\
    North Carolina State University\\
    Raleigh, North Carolina\\
    jyuan23@ncsu.edu, mpsingh@ncsu.edu
}

\begin{document}
\maketitle
\pagestyle{plain}
\thispagestyle{plain}

\begin{abstract}
Conversations among online users sometimes \emph{derail}, i.e., break down into personal attacks. Such derailment has a negative impact on the healthy growth of cyberspace communities. The ability to predict whether ongoing conversations are likely to derail could provide valuable real-time insight to interlocutors and moderators. Prior approaches predict conversation derailment retrospectively without the ability to forestall the derailment proactively. Some works attempt to make dynamic prediction as the conversation develops, but fail to incorporate multisource information, such as conversation structure and distance to derailment. 

We propose a hierarchical transformer-based framework that combines utterance-level and conversation-level information to capture fine-grained contextual semantics.  We propose a domain-adaptive pretraining objective to integrate conversational structure information and a multitask learning scheme to leverage the distance from each utterance to derailment. An evaluation of our framework on two conversation derailment datasets yields improvement over F1 score for the prediction of derailment. These results demonstrate the effectiveness of incorporating multisource information.
\end{abstract}

\section{Introduction}
Online social platforms provide great opportunities for users to constructively converse and collaboratively develop ideas \citep{hua-etal-2018-wikiconv}. However, antisocial behaviors such as personal attacks impede the building of healthy and thriving online communities \citep{Cheng2015AntisocialBI}.

Recently, NLP has been applied to address antisocial behavior. However, most previous research aims only at detecting antisocial behavior once the misconduct has occurred \citep{10.1145/3025453.3026018, 10.1145/3041021.3051106}. This post hoc identification limits the actions platform moderators can take. What most platform moderators do after the detection of antisocial behavior is to remove uncivil content or suspend the poster's account. However, the damage would have been done, and the involved parties would have been discouraged from participating in future conversations. Another practical problem is that uncivil content could potentially be overlooked by the moderator. 

\begin{figure}[t]
    \centering
    \includegraphics[width=1\columnwidth]{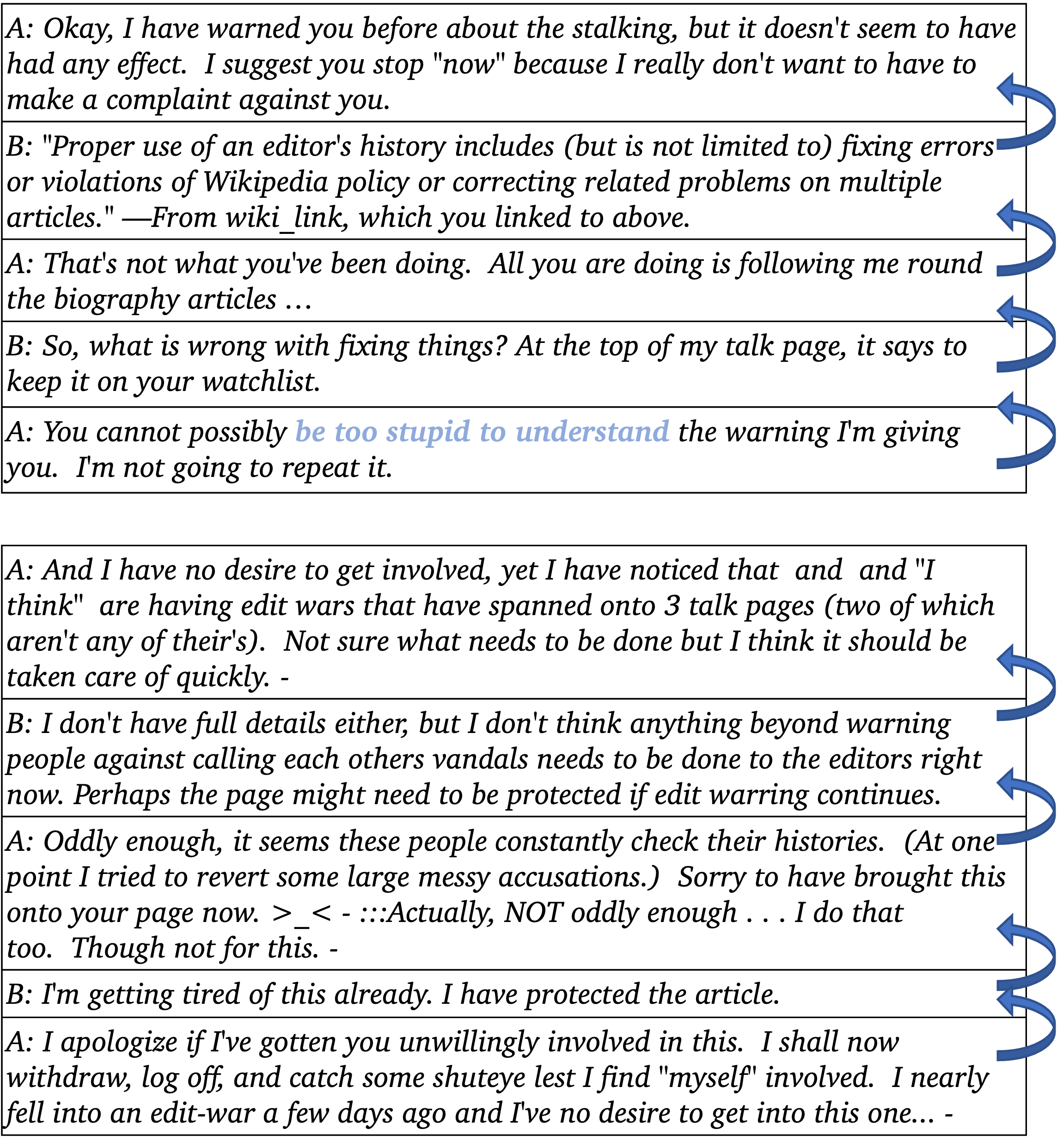}
    \caption{A pair of conversations with one developing into a personal attack in the end (top) and one staying civil throughout (bottom). Arrows show the ``reply\_to'' relationship between utterances.}
    \label{fig:an example of detrailing conversation}
\end{figure}

A more beneficial strategy would be to apply NLP to provide an early warning, possibly to interfere with the derailment as the conversation is developing. Figure~\ref{fig:an example of detrailing conversation} shows a pair of example conversations. Neither has personal attacks at the beginning, but one ends up with a personal attack (top) and another stays civil throughout (bottom). As an example intervention, the moderator may advocate for politeness or emphasize the rules of the platform when  potential derailment of an ongoing conversation is predicted. Achieving this strategy requires the model to learn and predict the dynamics of developing conversations, as opposed to the post hoc classification of past conversations. 

Forecasting conversation trajectory could enable use cases not limited to personal attacks. For example, the model could predict whether a conversation that persuades people to donate to charity will likely succeed or not \citep{wang-etal-2019-persuasion}, or whether a conversation is likely to change another person's view \citep{DBLP:conf/www/GuoZS20}. 

Conversation modeling and forecasting expose important challenges. First, there are complex dynamics on both the utterance and the conversation level. The semantics of the conversation is affected not only by the content of the utterances but also by the tree-like structure between utterances. The dynamics could change abruptly because of new utterances. Second, the number of utterances that will occur in a conversation is unknown. The conversation could stop at any time and a personal attack could happen at any moment as the conversation unfolds. An earlier warning is obviously better than a warning that comes up right before the attack but earlier warnings would have lower accuracy. When should the model make a prediction and how to trade off between timeliness aspect and accuracy? Third, conversation modeling challenges deep learning models because all utterances in a conversation need to be processed. Hence, the total length of a tokenized conversation produced by concatenating all utterances can be much longer than a single utterance, and can exceed the maximum input length limit (512) for BERT \citep{devlin-etal-2019-bert}. For instance, the average tokenized lengths for the two datasets we work with are 633 and 823, respectively.

Previous work addresses these challenges with different strategies. For example, \citet{hessel-lee-2019-somethings} mostly rely on hand-crafted features to model a conversation. \citet{chang-danescu-niculescu-mizil-2019-trouble} apply the LSTM architecture to capture the dynamics and consider only the first 80 tokens of each utterance.  Moreover, previous works \citep{chang-danescu-niculescu-mizil-2019-trouble, Janiszewski2021TimeAI} solely rely on textual semantics and disregard information such as conversational structure. 

To investigate conversation modeling for derailment prediction, we raise three interrelated research questions:
\begin{description}

    \item [RQ1] Is it effective to leverage pretrained language models for conversation modeling tasks and in what way?
    \item [RQ2] How can we leverage the information inherent in a conversation, such as distance from each utterance to the derailing utterance to augment the dataset?
    \item [RQ3] Does conversation structure matter for the derailment prediction and how do we integrate it into the model?

\end{description}

We answer these questions with a hierarchical transformer framework that tackles the above challenges by leveraging pretrained language model. We design ways to integrate various sources of information and explore how each component other than textual content contributes to the modeling of an ongoing conversation. Specifically, we propose a multitask training scheme to leverage the time factor for a conversation to derail, and a pretraining scheme to use conversational structure.

\section{Related Work}
\subsection{Antisocial Behavior in Cyberspace}
Previous research defines and detects various aspects of antisocial behaviors in online platforms. These antisocial behaviors, include toxicity \citep{pavlopoulos-etal-2020-toxicity, Ive2021RevisitingCT}, abusive language and content \citep{Vidgen2019ChallengesAF}, hate speech \citep{Fortuna2018ASO, Mozafari2019ABT}, trolling \citep{Mojica2017ATH}, offense \citep{Meaney2021SemEval2T},and racism \citep{field-etal-2021-survey}. Earlier work primarily relies on hand-crafted features \citep{hessel-lee-2019-somethings, zhang-etal-2018-conversations}, whereas recent work takes advantage of deep neural networks \citep{chang-danescu-niculescu-mizil-2019-trouble}. In contrast to detection of antisocial behaviors, another line of work \cite{conv-alright} takes a different perspective to study early cues and design metrics for quantifying and predicting
prosocial outcomes in online conversations.

Much of the past work focuses on classifying the type of antisocial behavior retrospectively with a single piece of text without considering the context. \citet{zhang-etal-2018-conversations} propose a task to predict whether an ongoing conversation will evolve into a personal attack as it develops. \citet{pavlopoulos-etal-2020-toxicity} detect and measure toxicity in context by considering the parent of a comment. Our work is a natural extension of previous work focusing on exploiting inherent contextual information to make fine-grained predictions of the trajectory of an ongoing conversation.
\subsection{Conversation Modeling}
Dialogue modeling is a promising line of research. \citet{Khanpour2016DialogueAC} address the classification of dialogue acts, which involves giving a predefined act type to each utterance. However, this type of classification focuses on utterance-level prediction. \citet{10.1145/2806416.2806493} propose a widely adapted architecture for conversation modeling, which applies a hierarchical recurrent neural network for encoding utterance and context, respectively. \citet{chang-danescu-niculescu-mizil-2019-trouble} leverage the same architecture and pretrain the model with domain data comprising over 1 million conversations. 

Our work differs from previous research in two aspects. First, we explore new ways of modeling conversation data by leveraging pretrained languages models. Second, we propose a new pretraining goal to incorporate the inherent tree structure of conversations into the model and evaluate its effectiveness.

\subsection{Domain Adaptive Pretraining}
Since the advent of BERT \citep{devlin-etal-2019-bert}, the pretraining-then-finetuning paradigm has been popular \citep{howard-ruder-2018-universal}. However, in some domain-specific tasks, this paradigm does not work well due to the lack of annotated data. \citet{gururangan-etal-2020-dont} propose another advanced computing paradigm, pretraining, domain-adaptive pretraining, and finetuning, to leverage the unlabeled domain-specific data. During domain adaptive pretraining, the training scheme is usually the same as during general pretraining, that is, using a masked language model. 

We follow \citepos{gururangan-etal-2020-dont} line of thinking but pretrain the model to identify the parent comment each utterance replies to.  

\section{Methodology} \label{ssec:3}
We now describe our model for evaluating and forecasting conversation development by integrating multisource information. We experiment with four settings, where additional information is consolidated incrementally during training to resolve each of our research question:

\begin{itemize}
    
    \item BERT. \citep{kementchedjhieva-sogaard-2021-dynamic} works on the same datasets with a simple BERT model. They concatenate all utterances and add a classification head on top of BERT. We follow the descriptions in their work to conduct the same experiment. 
    
    \item Hierarchical-Base. For \textbf{RQ1}, we leverage the pretrained language model to design a hierarchical transformer model that encodes the utterance-level and the conversation-level information, respectively. 
    
    \item Hierarchical-Multi. For \textbf{RQ2}, on top of Hierarchical-Base, we propose a multitask learning scheme and leverage the distance from each utterance to the derailment utterance as an auxiliary training objective. An intuition is that the distance till derailment can provide a fine-grained signal to the model.
    
    \item Hierarchical-Multi+Pretrain. For \textbf{RQ3}, we propose to use the inherent utterance structure, as captured by the ``reply-to'' attribute for each utterance. Each conversation is viewed as a tree structure. We set up a pretraining objective to predict the parent of each utterance.
\end{itemize}

\subsection{Problem Formulation}

We now define the problem formally. A conversation is a sequence of utterances, $C = \{u_1, \dots, u_n\}$, where $n$ is the number of utterances in the conversation. Each utterance consists of a sequence of words, $u = \{w_1, \dots, w_m\}$, where $m$ is the number of words in the utterance. Each conversation comes with a label $d = positive$ or $d=negative$, where $positive$ denotes there is a personal attack (derailment) at utterance $u_n$, and $negative$ denotes the conversation is civil throughout. A data sample can be represented as a tuple $(\{u_1, \dots, u_n\}, d)$. 

We focus on predicting the possibilities of derailment for \textit{ongoing} and \textit{civil} conversations, i.e., how likely a civil conversation is to lead to a personal attack as it develops. To this end, exchanges between speakers after the first derailment are ignored when preparing the dataset so that the model is fed with civil utterances during training and inference. Therefore, positive samples are in the format of $\{0, \dots, 0, 1\}$, and negative samples are in the format of $\{0, \dots, 0, 0\}$, whereas $1$ represents personal attack utterance and $0$ represents civil utterance.

We don't consider what happens after a personal attack, and defer it to future research. The evaluation process is dynamic, which means that for each conversation the model makes sequential predictions for a list of inputs $(\{u_1\}, \{u_1, u_2\}, \dots, \{u_1, u_2, \dots, u_{n-1}\})$. The model stops whenever a positive prediction is made, which indicates that the conversation will derail. Naturally, for a conversation with label $negative$, the model should make $n-1$ negative predictions.

The four model settings can be categorized into two types. The first one is a plain BERT model that concatenates all utterances and doesn't differentiate between utterance-level and conversation-level encoding. The other three models are variants of a hierarchical transform, which encode each utterance first, and then apply attention layers to capture conversation-level information. Below is a detailed description for the architecture.
	
\subsection{Utterance-Level Encoder}

For each utterance $u = \{w_1, \dots, w_m\}$, we leverage the pretrained language model to capture rich semantics. Specifically, we use a transformer-based model with the same configuration as RoBERTa-base \citep{Liu2019RoBERTaAR}, initialized with  pretrained weights from Huggingface\footnote{https://huggingface.co/}. RoBERTa improves over BERT by employing dynamic masking with ten times as much training data. We follow the preprocessing steps to tokenize the utterance and append special tokens [CLS] at the front and [SEP] at the end. Before feeding the token embedding into the first-layer transformer, we add a pretrained positional embedding to each token. The maximum input length for RoBERTa is 512 and we cut off extra tokens if the tokenized utterance length exceeds the limit. Finally, we take the embedding of the special token [CLS] from the last layer's output as utterance representation.
    
\subsection{Conversation-Level Encoder}
Derailment should not be considered a singular attribute of an utterance as it is the result of an entire conversation. Therefore, we consider the cumulative effect of previous utterances. For each conversation $C = \{u_1, \dots, u_n\}$, we obtain the utterance embedding $E$ from the first-level transformer, and then feed the sequence of utterance embedding to a few identical transformer layers. Similarly, we use the [CLS] embedding from the last layer as a representation of the entire conversation and feed it to a classifier. The classifier is made up of one fully connected linear layer for the binary classification head. We choose transformer layers over an LSTM layer because multihead attention mechanisms have an edge over the traditional LSTM model. To reduce computational cost, instead of applying a full transformer model, we use only four transformer layers. 

\begin{figure}[t]
\centering
\includegraphics[width=1\columnwidth]{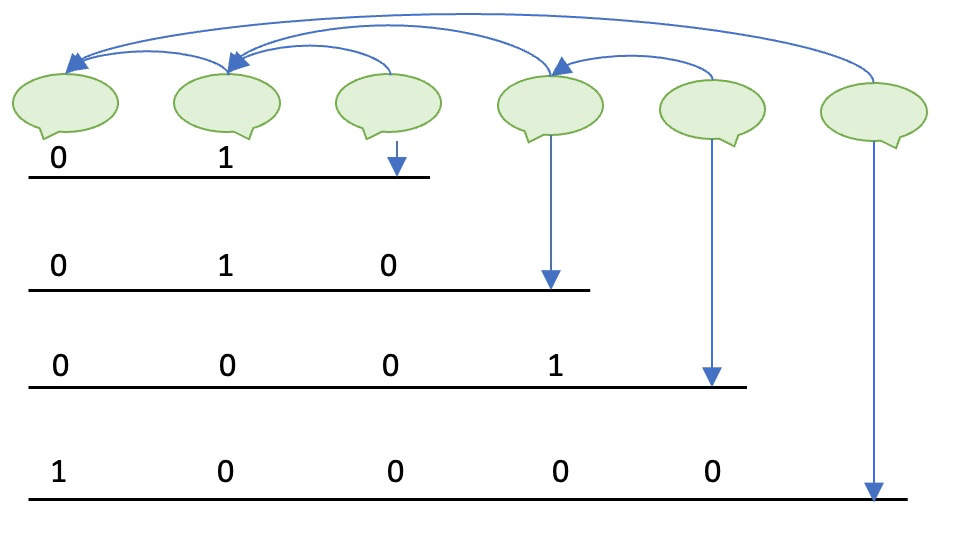}
\caption{Illustration of structure pretraining. Ellipses denote an utterance sequence with the ``reply\_to'' relation between them. The labels represent the ground truth for each corresponding subsequence.}
\label{fig:pretrain}
\end{figure}

\subsection{Multitask Training with Distance to Derailment} \label{ssec:3.4}

Following previous work \citet{chang-danescu-niculescu-mizil-2019-trouble}, we simulate how conversations evolve in reality and predict conversation derailment dynamically. At inference time, the model is fed with the sequence $\{u_1, u_2, \dots, u_{n-1}\}$ utterance by utterance and makes a prediction at each step. The model stops whenever a positive prediction is made, which indicates that the conversation is deemed to derail. At training time, however, \citet{chang-danescu-niculescu-mizil-2019-trouble} apply a static training strategy, where the model is trained only with full sequences up to the derailing utterance $\{u_1, u_2, \dots, u_{n-1}\}$. We posit that this discrepancy may bias the model to overestimate the probability of derailment for long inputs. Therefore, we propose to unify training and inference with the same dynamic strategy by adopting a regression task besides the binary classification task.

Consider a sample pair $(\{u_1, u_2, u_3, u_4\}, positive)$, $(\{u'_1, u'_2, u'_3, u'_4\}, negative)$. We observe that the distance from each civil utterance to the derailing utterance could provide additional cues for the model to learn. By predicting the distance to derail, another benefit is that we can expand the training set by a factor of the average conversation length. 

For the positive sample, we can train on $(\{u_1\}, 3)$, $(\{u_1, u_2\}, 2)$, $(\{u_1, u_2, u_3\}, 1)$ for the regression task, where the targets $3, 2, 1$ represent the distance from the current sequence to the derailing utterance $u_4$. In the extreme, if the target is infinity, it means the conversation doesn't derail. In other words, a larger target implies a lower chance of derailment. 

\begin{figure}[t]
\centering
\includegraphics[width=1\columnwidth]{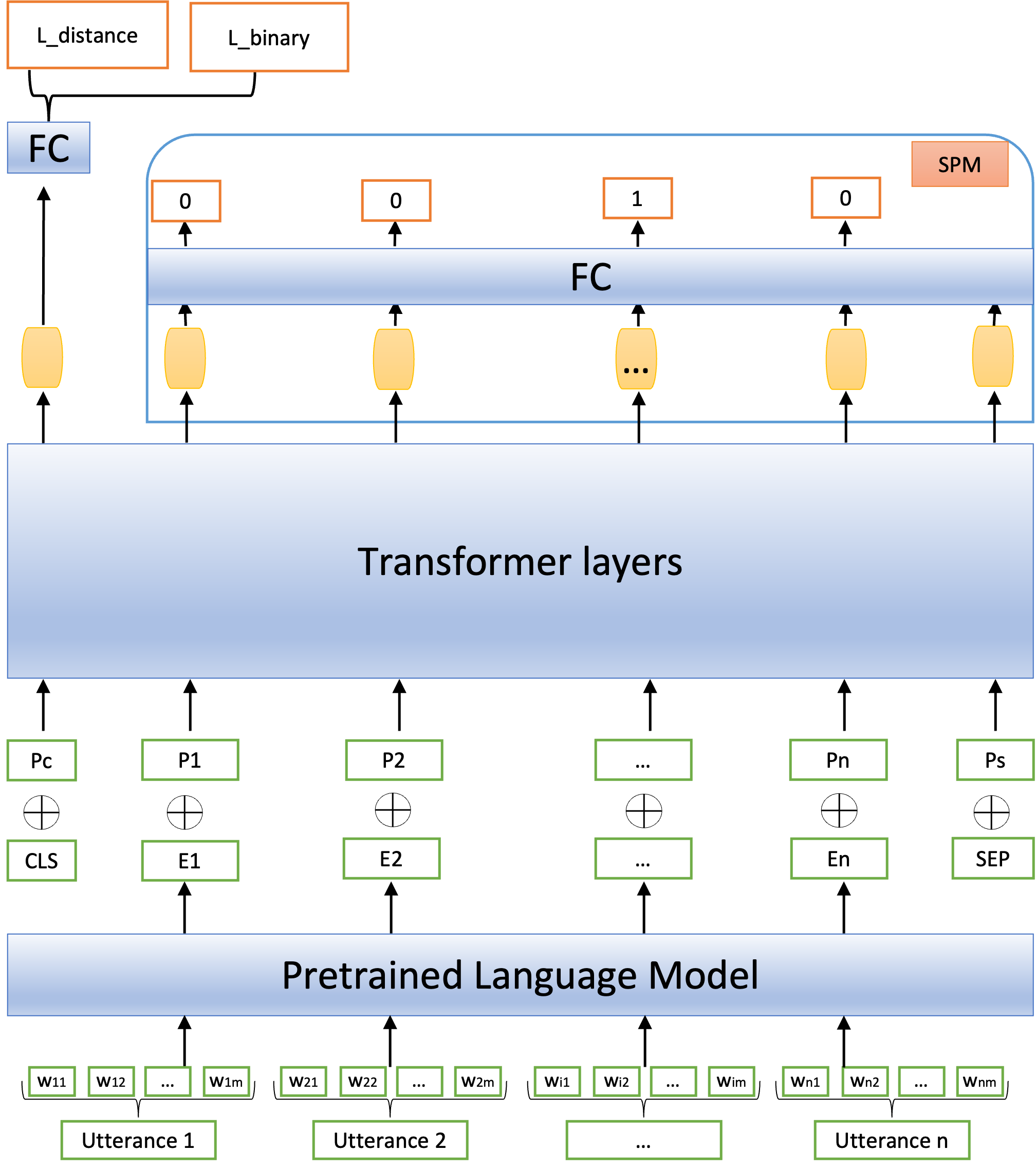}
\caption{Illustration of a hierarchical transformer with a multitask learning scheme. The top-left component represents the multitask training scheme. SPM is the structural pretraining module.}
\label{fig:overall architecture}
\end{figure}

Therefore, for the negative sample $(\{u'_1, u'_2, u'_3, u'_4\}, 0)$, we expand into $(\{u'_1\}, \infty)$, $(\{u'_1, u'_2\}, \infty)$, $(\{u'_1, u'_2, u'_3\}, \infty)$, $(\{u'_1, u'_2, u'_3, u'_4\}, \infty)$. A practical consideration is that if we set the target to be infinity for negative samples, the regression loss would be too large. Practically, we can simply set the longest length of conversation in the dataset as the target for each expanded negative sample. We add another regression head following the fully connected linear layer. Figure~\ref{fig:overall architecture} illustrates the multitask training architecture on the top left. Our new loss function becomes $L = \alpha L_{distance} + (1-\alpha)L_{binary}$, where $\alpha$ controls the weight of each loss. We use mean squared error loss for $L_{distance}$ and cross entropy for $L_{binary}$. 

\subsection{Conversation Structure Pretraining}

Conversations between a group of people on social media  usually come with an inherent tree structure. Platforms such as Reddit, Wikipedia talk page, and Twitter have a clearly defined ``reply-to'' attribute for each comment. Previous research focuses less on the structure when modeling  conversations. To investigate whether such structure is relevant in conversation development, we propose a scheme to pretrain our model on such structure in an unsupervised way. Specifically, we feed a sequence of utterance embeddings $\{E_1, E_2,\dots, E_n\}$ to the second-level transformer model as usual, and denote the output from the last transformer layer as $\{O_1, O_2,\dots, O_n\}$, which is followed by a fully connected layer with a softmax activation function. 
\[O_i = \text{Trans\_Layer}(E_i), i = 1, \dots, n-1\]

\[p_i = \softmax{(\text{FC}(O_i))}, i = 1, \dots, n-1\]

and the loss function would be 

\[ L = -\sum_{i=1}^{n-1}{y_i\log{p_i}}\]
where $y_i=1$ if the $ith$ utterance is the parent of the $nth$ utterance. 

Figure~\ref{fig:pretrain} illustrates the ground truth labels for all the subsequences of a conversation.  We refer to the pretraining module as the structural pretraining module (SPM).

\subsection{Data Augmentation Strategy}
\label{ssec:3.6}

One advantage of multitask training and conversation structure pretraining is that the dataset can be augmented by a factor of the average length of the conversations, as explained in Section~\ref{ssec:3.4}. This straightforward strategy, however, may not produce the best results. After examining some conversations, we observe that for the longer conversations, the first few utterances have weak indications of derailment, which may confuse the model. We assume that the derailment is the result of the cumulative effect of multiple exchanges. Therefore, we apply a different augmentation strategy to keep at least half of the previous utterance. Suppose the length of the conversation is $n$, we start expanding the data at length $i=\lfloor n/2 \rfloor$. Now each data sample is extended to a sequence of samples $\{u_1,\dots, u_{i}\}, \{u_1,\dots, u_{i}, u_{i+1}\}, \dots,\{u_1,\dots, u_{i},$ $ u_{i+1}, \dots, u_{n-1}\}$. Our results show that this strategy works better. 

\section{Experiments and Results}
We evaluate our model on two canonical conversation derailment datasets proposed in previous research \citep{zhang-etal-2018-conversations} \citep{chang-danescu-niculescu-mizil-2019-trouble}. Both datasets reflect the same philosophy with respect to collecting the data. The essential idea in both is to construct (1) positive samples, where the first few utterances are civil but eventually develop into a personal attack and (2) negative samples, where the whole conversation is civil. Another consideration is to avoid potential bias related to conversation topic. For example, if the topic distribution between negative and positive samples is different, the model is likely to capture that and can predict high probability of derailment for conversations that share the similar topics with the positive samples. One way to counter this challenge is to make sure that every positive sample has a corresponding negative sample where the conversations cover similar topics.  Therefore, the procedure has two steps. First, identify conversations that contain personal attacks. Second, for each derailing conversation, collect a civil conversation that covers the same topic. As a result of this procedure, the dataset is balanced between derailing and civil conversations. 

\subsection{Datasets}

\paragraph{Wikipedia talk page (WTP)}
The WTP dataset was introduced by \citep{zhang-etal-2018-conversations} and then expanded by \citep{chang-danescu-niculescu-mizil-2019-trouble} using the same procedure. Every Wikipedia article is associated with a Talk Page, where Wikipedia editors discuss its editing. Each page usually has multiple sections focusing on the discussion of different editing problems. Every section is in the form of a conversation. The goal is to select conversations that start out as civil but derail into personal attacks afterwards. Wikipedia contains millions of pages and conversations. To alleviate the effort of manually going through all conversations, a toxicity classifier is used to select candidate conversations that contain toxic utterances. Toxicity, however, is not always equivalent to personal attacks. Therefore, after selecting the candidates, a manual screening is further applied to select conversations with  personal attacks.

The classifier used is provided by Perspective API, which is trained on Wikipedia talk page comments that have been annotated by humans. The classifier provides a toxicity score ranging from 0 to 1 for each utterance. Two types of conversation are preselected: (a) those that are civil throughout---all comments in the conversation have a toxicity score below 0.4 and (b) those that are civil for the first exchange (two comments), but turn toxic afterwards---there is a comment with toxicity score above 0.6. These two numbers are chosen empirically by running examples with the API. When the score is lower than 0.4, it has high fidelity that it is civil and when it is over 0.6, it is toxic. However, the exact numbers do not matter because the classifier is used only as a first step filtering of candidate conversations. As explained above, to avoid the model from capturing spurious correlations, such as conversation topics or length, each positive sample (conversations starting out civil and ending with personal attacks, i.e., the candidates from the first step) is paired with a negative sample (conversations are civil throughout). These negative samples selected as follows: (1) the are from the same Wikipedia talk page so their discussions are about the same article; (2) they have similar lengths; and (3) they take place close in time. We refer the reader to Section~3 of the original paper \cite{zhang-etal-2018-conversations} for additional detail. This procedure produces a dataset that contains \numprint{2,094} pairs of conversations, splitting into \numprint{60}--\numprint{20}--\numprint{20} segments. Formally, we denote 1 as a personal attack comment and 0 as a civil comment. Then, a sample data pair can be represented as ${(0,0,0,0), (0,0,0,0,1)}$. As a result of the procedure, the dataset has the same number of positive and negative samples.

\paragraph{Reddit ChangeMyView (CMV)}
The Reddit CMV dataset was crafted by \citep{chang-danescu-niculescu-mizil-2019-trouble}. ChangeMyView is a subreddit where reddit users post their opinions and challenge other users to change their views. There is a specific rule stating ``Rule 2: Don't be rude or hostile to other users.'' The platform moderator may delete any comment that appears to have personal attacks and replace the comment with the word ``deleted.'' Then, all previous comments up to the ``deleted'' comment make up a positive conversation sample with derailment at the end. The deleted comment is not visible. However, as the setting is to predict whether a conversation will derail or not, rather than to classify an existing derailing utterance, the content of the derailing utterance is not needed for training. To apply the topic and length control pairing, as in the WTP datasets, each positive and negative pair of similar length is chosen from the same top-level post. An extra control is also applied to select conversations where the deleted comment is from a user who previously participated in the conversation. We refer
the reader to section 3 of the original paper \citep{chang-danescu-niculescu-mizil-2019-trouble} for more details. This procedure produces a dataset of \numprint{3,421} pairs of conversations, which is also split into \numprint{60}--\numprint{20}--\numprint{20} parts. One thing to note is that there is no post hoc annotation and examination of this dataset. Therefore, there is no guarantee that for either positive or negative conversations, all comments prior to the last one would be civil. Thus, the CMV dataset may contain more noise than the WTP dataset. 

\begin{table*}[t]
\centering
\begin{tabular}{l@{~} c c c c c c c c c c}\toprule
& \multicolumn{5}{c}{Wikipedia talk pages} & \multicolumn{5}{c}{Reddit CMV} \\\cmidrule(lr){2-6} \cmidrule(lr){7-11}
  Model & Accuracy & Precision & Recall & FPR & F1 & Accuracy & Precision & Recall & FPR & F1\\ %[0.5ex] 
 \midrule
 BoW & 56.5 & 55.6 & 65.5 & 52.4 & 60.1 & 52.1 & 51.8 & 61.3 & 57.0 & 56.1 \\
 % CRAFT & 60.6 & 57.3 & \textbf{77.6} & 57.4 & 66.0 & 57.4 & 52.2 & 57.2 & 52.3 & 54.6 \\
  CRAFT & 64.4 & 62.7 & 71.7 & - & 66.9 & 60.5 & 57.5 & \textbf{81.3} & - & 67.3 \\
 HRED & 63.9 & 63.8 & 64.1 & \textbf{36.2} & 64.0 & 55.6 & 54.6 & 65.8 & 54.7 & 59.7 \\
 %\midrule
  BERT & 63.8 & 61.1 & 75.7 & 48.1 & 67.6 & 65.7 & 64.4 & 70.2 & 38.8 & 67.1\\
 Hierarchical-Base & 62.9 & 60.3 & 75.2 & 49.5 & 66.9 & 64.3 & \textbf{67.1} & 56.2 & \textbf{27.6} & 61.1 \\
 Hierarchical-Multi& 65.2 & 62.3 & \textbf{76.9} & 46.4 & \textbf{68.9} & 64.2 & 62.0 & 73.8 & 45.2 & \textbf{67.4} \\
 Hierarchical-Multi+Pretrain& \textbf{65.2} & \textbf{64.2} & 69.1 & 38.6 & 66.5 & \textbf{66.2} & 66.5 & 65.2 & 32.8 & 65.9 \\ 
 \bottomrule

\end{tabular}
\caption{
Results of the proposed model on two datasets, compared to three previous approaches (BoW and CRAFT \citep{chang-danescu-niculescu-mizil-2019-trouble}, HRED \citep{Janiszewski2021TimeAI}), BERT\citep{kementchedjhieva-sogaard-2021-dynamic}. Hierarchical-Base (base version of the hierarchical transformer), Hierarchical-Multi (hierarchical transformer with multiple learning scheme), Hierarchical-Multi+Pretrain (hierarchical transformer with multiple learning scheme and pretraining module). }
\label{table:results}
\end{table*}

\subsection{Experimental Setup}
\paragraph{Baseline} We compare our approach with a few different works from previous research. Specifically, We compare the performance with a straightforward bag-of-words model, which simply concatenates all utterances and converts them into a bag-of-words vector. We also compare with the CRAFT model proposed by the original paper \cite{chang-danescu-niculescu-mizil-2019-trouble}. There are two recent works from \cite{Janiszewski2021TimeAI} and \cite{kementchedjhieva-sogaard-2021-dynamic}, which evaluate on the same dataset. 

\paragraph{Training process} To facilitate the learning process, we experimented with different configurations. We implemented our model with the Hugging Face library and set up the learning rate to be 1e-5 with a batch size of 32 for the multitask learning part. We experiment with different $\alpha$ values and set it to be 0.3. The experiment shows that we need to be conservative with the regression task. The model was trained with the Adam optimization algorithm. We observe that using only the first or second utterance as inputs would have a negative impact on performance, possibly due to the fact that the first and second utterances do not provide enough cues for the model to predict derailment. Therefore, we adopt the strategy described in Section~\ref{ssec:3.6}. With this strategy, the target for regression task is in the range of 0 to 5. During training, we evaluate the performance on the dev split every 100 iterations and only save the checkpoint if the performance of the current iteration is better than that of the last checkpoint. For the SPM adaptive pretraining module, as the goal of the training is to enable the model to be able to capture general structural information existing in the conversation, we adopt the full data augmentation strategy as described in Section~\ref{ssec:3.4}. 

\paragraph{Evaluation}
Following \citet{Janiszewski2021TimeAI}, we evaluated the performance of the model with accuracy, precision, recall, FPR (false positive rate), and F1 score. To have a fair comparison, during evaluation, we looked only at the outcome from the binary prediction head and ignored the prediction from the distance-to-break head. The evaluation was done in a progressive manner. That is, for a conversation of length L, whenever the model makes a positive prediction when feeding with utterance sequence $(\{u_1\}, \{u_1, u_2\}, \dots, \{u_1, u_2, \dots, u_{n-1}\})$, it counts as a derailment prediction. If the entire preceding sequence is predicted as negative, the conversation is deemed to be not derailed. 

\subsection{Results and Analysis}

Table~\ref{table:results} shows the results of our various model variants, together with the models from previous research. One thing to note is that the performance of CRAFT in \cite{Janiszewski2021TimeAI} and \cite{kementchedjhieva-sogaard-2021-dynamic} are a bit lower than what the original paper reported, due to variations during training. We cite the results from \cite{kementchedjhieva-sogaard-2021-dynamic} which has taken the average of ten runs of CRAFT. The BERT model in the table is the architecture from \cite{kementchedjhieva-sogaard-2021-dynamic}, which simply add a classification head on top of the original BERT model. We follow the static training details described by the paper and report the results we have. 
Hierarchical-Base is our base model of the hierarchical transformer, which consists of only one binary prediction head without data augmentation. Hierarchical-Multi is the multitask learning model, which has two prediction heads and uses data augmentation during training. Hierarchical-Multi+Pretrain includes pretraining with structural information followed by the Hierarchical-Multi model. Overall, our models achieve better performance regarding most of the metrics. Between our own variants, the result are mixed. 

\begin{figure}[t]
\centering
\includegraphics[width=1\columnwidth]{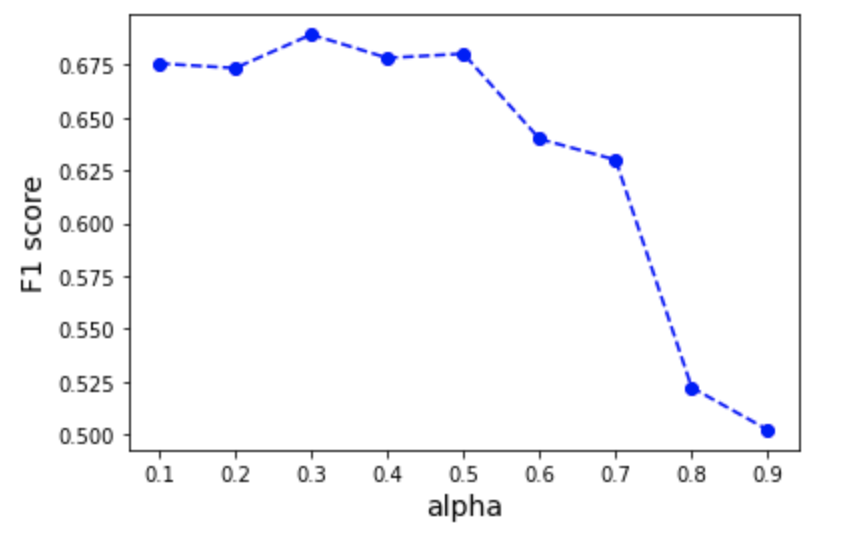}
\caption{F1 score with respect to different $\alpha$ value on the WTP dataset}
\label{fig:alpha}
\end{figure}

We also experiment with different values of $\alpha$ with Hierarchical-Multi on the WTP dataset to investigate the optimal weight of each task contributing to the loss. Fig \ref{fig:alpha} shows that the optimal $\alpha$ should be 0.3. The insight is that we should be fairly conservative with respect to the regression task. This observation echoes the fact that we should not augment the data to the full length of each conversation. The model learns better when more utterances are being observed.

A major difference between our architecture and previous models is that we leverage the power of pretrained language models and self-attention mechanisms. Both CRAFT \citep{chang-danescu-niculescu-mizil-2019-trouble} and HRED \citep{Janiszewski2021TimeAI} consist of two LSTM layers, and they pretrained the model over 1 million conversations with an autoregressive language model objective to get the best performance, which is a heavy cost and requires gathering a large amount of data from the same domain. It's also not easy to adapt the pretrained language model to other tasks.  As we can see, all model variants based on pretrained language models achieve better F1 scores on both the WTP and CMV datasets. HRED has the lowest FPR on the WTP dataset while CRAFT has the highest recall on the same dataset. For all other metrics on both datasets, our models have  higher performance. The BERT model is from \cite{kementchedjhieva-sogaard-2021-dynamic}, which has a classification head on top of BERT. Even though with a simple BERT model, the performance is still competitive. Therefore, we answer \textbf{RQ1} postively. 

Comparing the three different variants of our hierarchical model, the multitask learning architecture yields the best F1 scores among all models on both datasets, which demonstrates the effectiveness of our multitask learning and data augmentation strategy. One thing to notice from our experiment is that, if we change our data augmentation strategy from ratio sampling to full sampling, the performance decreases, which indicates that the model needs more context to learn the factors leading to derailment. Conversely, the base version of hierarchical transformer has a lower FPR than Hierarchical-Multi on the CMV dataset, which implies that the distance-to-derailment information encourages the model to make positive predictions. This is echoed by the fact that the Hierarchical-Multi model has a higher recall rate. Hierarchical-Multi has the highest F1 score on the WTP dataset. Both Hierarchical-Multi and Hierarchical-Multi+Pretrain have higher performance than Hierarchical-Base in terms of F1 score on the CMV datasets, which demonstrates the effectiveness of the multitask learning and data augmentation strategy. Therefore, we answer \textbf{RQ2} positively. 

Contrary to our expectations, adding the reply-to pretraining process does not yield improved performance. There might be three reasons for this. First, we didn't pretrain our model on other data within the same domain, but  limited our pretraining to within the two datasets. Although the dataset is augmented by a factor of conversation length, the total number of data points is far less than the amount of data that is used by \citet{chang-danescu-niculescu-mizil-2019-trouble}. Second, the reply-to relation between the utterance in these two datasets possibly doesn't align well with the semantics of the utterances. Third, derailment may not relate too much to the semantics of each utterance. We observe that most derailments happen due to impoliteness. The conversation topics in both datasets are diverse and the pretraining may cause the model to capture some noises in the dataset. At the end, we answer \textbf{RQ3} negatively. 

\subsection{How Early is the Warning?}

\begin{table}[t]
\centering
\begin{tabular}{l@{~} c c}\toprule
  & WTP & CMV\\
   \midrule
 BERT & 2.73 & 3.98 \\
 Hierarchical-Base & 2.96 & 3.94\\
 Hierarchical-Multi& 2.86 & 3.45\\
 Hierarchical-Multi+Pretrain& 2.98 & 3.78 \\ 
 \bottomrule

\end{tabular}
\caption{Mean number of utterances between the issue of warning and the derailment utterance. BERT\citep{kementchedjhieva-sogaard-2021-dynamic}. Hierarchical-Base (base version of the hierarchical transformer), Hierarchical-Multi (hierarchical transformer with multiple learning scheme), Hierarchical-Multi+Pretrain (hierarchical transformer with multiple learning scheme and pretraining module). }
\label{table:horizon}
\end{table}

\begin{figure}[t]
    \centering
    \includegraphics[width=0.8\columnwidth]{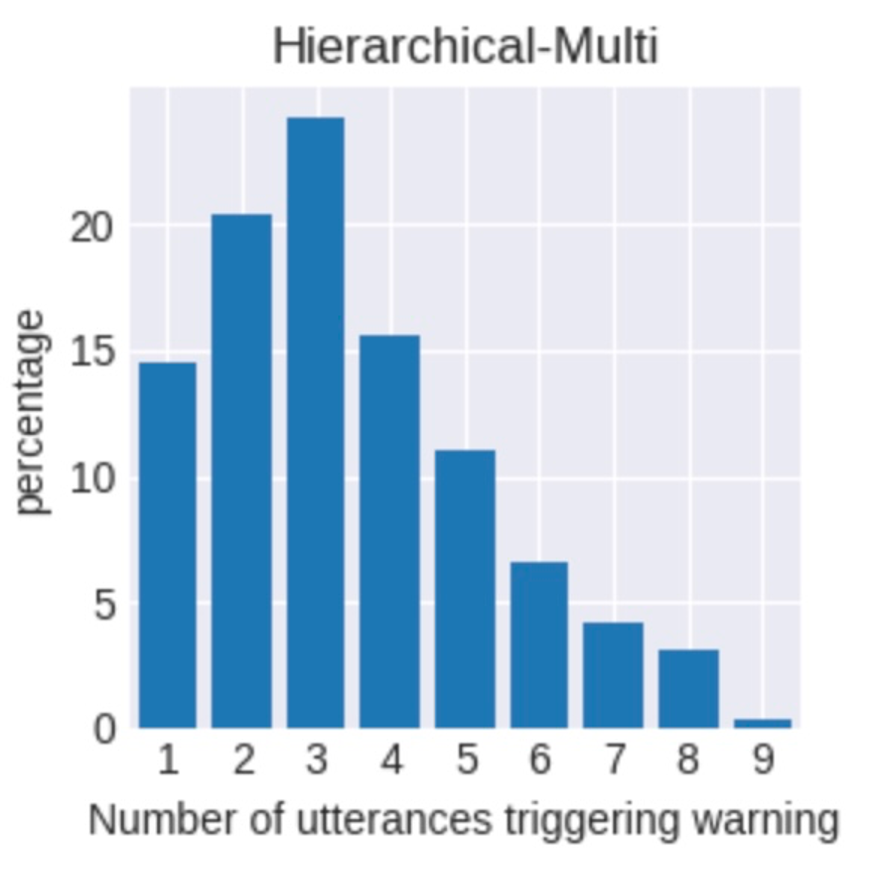}
    \caption{Percentage of the number of utterances elapsed when the model makes a positive prediction of derailment.}
    \label{fig:early warning}
\end{figure}

To investigate how early the warning is triggered for a conversation, we perform an analysis on the two datasets. Figure~\ref{fig:early warning} shows the distribution of the percentage of utterances that have elapsed when the Hierarchical-Multi model makes a prediction of derailment for all the positive samples in the testing set. We observe that around 80\% of warnings are issued when fewer than five utterances have been seen by the model and the average length of the derailing conversations in the testing set is 7.2.  We also show the average number of utterances between the issuance of a warning and the derailment utterance for four models. The distance is around 3 for the WTP dataset and 4 for the CMV dataset. These figures generally match with those in previous research \cite{kementchedjhieva-sogaard-2021-dynamic}, \cite{chang-danescu-niculescu-mizil-2019-trouble} have discovered. 

\section{Conclusion and Future Work}

We focus on a new intervention perspective regarding detecting and moderating toxic and abusive behaviors in online forum conversations. Rather than predicting whether a conversation contains toxic content retrospectively, we seek to predict whether an ongoing conversation will break down. We propose a hierarchical transformer architecture to capture both utterance-level and conversation-level semantics leveraging the power of the pretrained language models. In addition, we propose new ways to integrate conversational structure and the distance-to-derailment information and achieve better F1 scores than previous approaches on two canonical conversation derailment datasets.

Although our model mainly addresses the problem of predicting conversation derailment dynamically, it is a general approach for conversation modeling and can be adapted to address other conversation prediction tasks. For example, to predict whether a goal-oriented conversation would succeed in the end, we can set the label for successful conversations as $1$ and unsuccessful conversation as $0$. In addition to a simple binary prediction, our distance-to-derailment prediction could provide extra time-sensitive information. For tasks where the inherent conversation structure matters, our model provides a natural way to exploit such structure information. We defer the application of our approach on other conversation prediction tasks to future work. 

With all the strengths that come with our approach, we identify some limitations during our experiments. The first limitation is that we have limited our reply-to structure pretraining to the two datasets we evaluate, which doesn't seem to provide enough data for this kind of pretraining. Therefore, to leverage the structure information, we could continue to train the model on any conversations that happened within the same forum. Another option is to adopt specialized architectures that align well with structural data, such as graph neural networks, which are specifically designed for capturing tree-like structure information. A second promising direction is to capture speaker identity. For most conversations, there might be a pattern where certain types of users are more inclined to attack other people. The ability to model different speaker types may be important in such kinds of prediction. 

\section{Ethical Discussion}
Our work proposes ways to model conversation and predict whether the conversation will develop into personal attacks. The datasets we work on contain only user IDs from Reddit and Wikipedia talk page, which do not reveal personal identities and pose no privacy issues. The proposed framework and trained model can be applied by social platforms to assist with content moderation, where early warnings can be issued to prevent personal attacks from happening. As with many other pretrained deep learning models, our model could be exploited by users so that they learn the pattern to avoid the censor. Our model could also be limited in its ability to accurately capture conversation dynamics as the domain and topics evolve over time. However, we posit that active and continuous learning could help mitigate this problem. 
\bibliography{Jiaqing}

\end{document}